\documentclass[lettersize,journal]{IEEEtran}
\usepackage{amsmath,amsfonts}
\usepackage{algorithmic}
\usepackage{geometry}
\usepackage{algorithm}
\usepackage{xcolor} 
\usepackage{array}
\usepackage[caption=false,font=normalsize,labelfont=sf,textfont=sf]{subfig}
\usepackage{textcomp}
\usepackage{stfloats}
\usepackage{url}
\usepackage[compatibility=false]{caption}
\usepackage{verbatim}
\usepackage{graphicx}
\usepackage{booktabs}   
\usepackage{multirow}   
\usepackage{tabularx}
\usepackage{makecell}
\usepackage{subcaption}
\usepackage{booktabs}
\usepackage{multirow}
\usepackage{hyperref}
\usepackage{cite}
\begin{document}

\title{DENet: Dual-Path Edge Network with Global-Local Attention for Infrared Small Target Detection}

\author{Jiayi Zuo, Songwei Pei, Qian Li, Yuanzhuo Huang and Shangguang Wang
\thanks{Manuscript received (DATE); revised(DATE),  (DATE) and   (DATE); accepted (DATE). Date of publication (DATE); date of current version (DATE). \textit{(Corresponding author: Songwei Pei.)}}     
\thanks{The authors are with the School of Computer Science (National Pilot Software Engineering School), Beijing University of Posts and Telecommunications, Beijing 100876, China (e-mail: peisongwei@bupt.edu.cn).}
\thanks{Color versions of one or more of the figures in this letter are available online at http://ieeexplore.ieee.org.}

\thanks{Digital Object Identifier}
}
\markboth{Journal of \LaTeX\ Class Files,~Vol.~14, No.~8, August~2021}%
{Shell \MakeLowercase{\textit{et al.}}: A Sample Article Using IEEEtran.cls for IEEE Journals}


\maketitle

\begin{abstract}
Infrared small target detection (IRSTD) is crucial for remote sensing applications like disaster warning and maritime surveillance. However, due to the lack of distinctive texture and morphological features, infrared small targets are highly susceptible to blending into cluttered and noisy backgrounds. A fundamental challenge in designing deep models for this task lies in the inherent conflict between capturing high-resolution spatial details for minute targets and extracting robust semantic context for larger targets, often leading to feature misalignment and suboptimal performance. Existing methods often rely on fixed gradient operators (e.g., Sobel, Canny) or simplistic attention mechanisms, which are inadequate for accurately extracting target edges under low contrast and high noise. In this paper, we propose a novel Dual-Path Edge Network (DENet) that explicitly addresses this challenge by decoupling edge enhancement and semantic modeling into two complementary processing paths. The first path employs a Bidirectional Interaction Module (BIM), which uses both Local Self-Attention and Global Self-Attention to capture multi-scale local and global feature dependencies. The global attention mechanism, based on a Transformer architecture, integrates long-range semantic relationships and contextual information, ensuring robust scene understanding. The second path introduces the Multi-Edge Refiner (Multi-ER), which enhances fine-grained edge details using cascaded Taylor finite difference operators at multiple scales. This mathematical approach, along with an attention-driven gating mechanism, enables precise edge localization and feature enhancement for targets of varying sizes, while effectively suppressing noise. Extensive experiments on the IRSTD-1K and NUDT-SIRST benchmark demonstrate that DENet significantly outperforms state-of-the-art methods, achieving superior Mean Intersection over Union (mIoU) and pixel-level accuracy, while maintaining lower false alarm rates. Our method provides a promising solution for precise infrared small target detection and localization, combining structural semantics and edge refinement in a unified framework.

\end{abstract}


\section{Introduction}

\IEEEPARstart{I}{nfrared} small target detection (IRSTD)~\cite{b1,b46} is vital for remote sensing applications such as disaster rescue, maritime surveillance, and early warning systems~\cite{b65, b66, b78, b83, b84}. These mission critical scenarios require high detection reliability, as missed targets can cause severe risks to safety and infrastructure~\cite{b47,b48,b49,b50,b51}. Compared with visible or radar imaging, infrared sensors passively capture thermal radiation, which enables all weather operation and improves robustness under adverse conditions.~\cite{b70} Their long range sensing capability and fine resolution make them indispensable for identifying distant threats in both civilian and defense contexts.

\begin{figure}[t]
	\centering
	\includegraphics[width=\columnwidth]{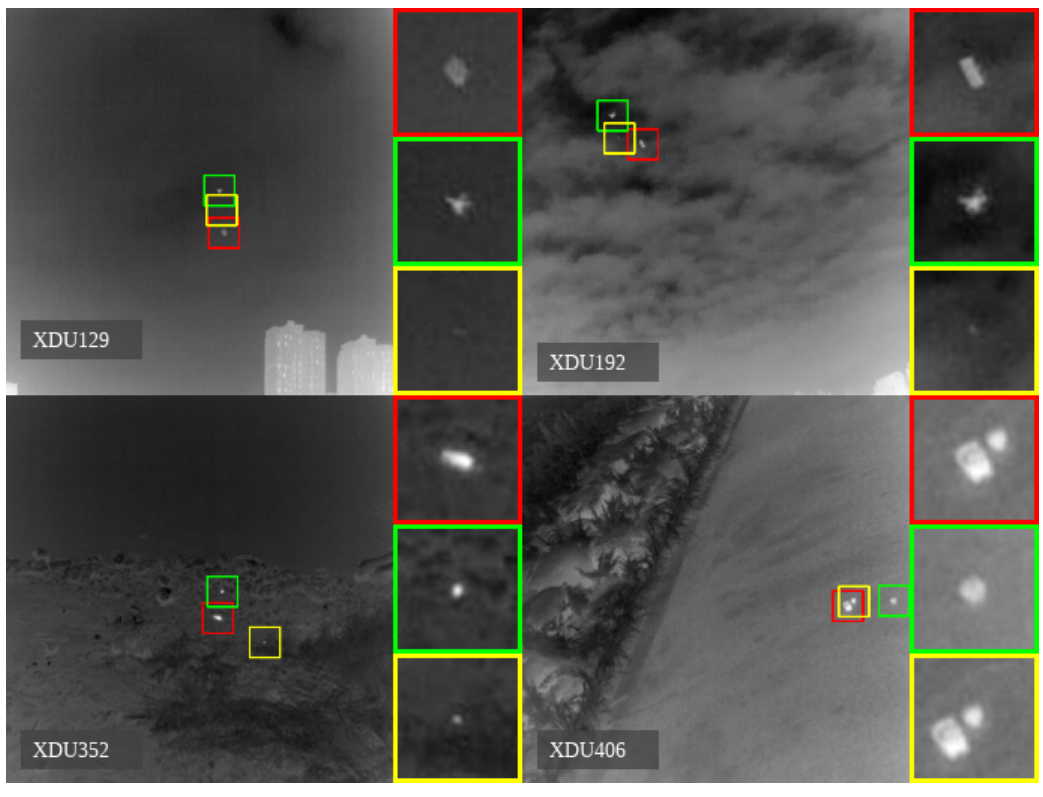}
	\caption{Target detection results illustrated across different infrared scenes with varying target sizes and complexities. The red bounding boxes indicate the largest targets, green bounding boxes mark the second largest targets, and yellow bounding boxes highlight the smallest targets in each scene.}
    \label{fig_big}
\end{figure}

\begin{figure}[t]
	\centering
	\includegraphics[width=\columnwidth]{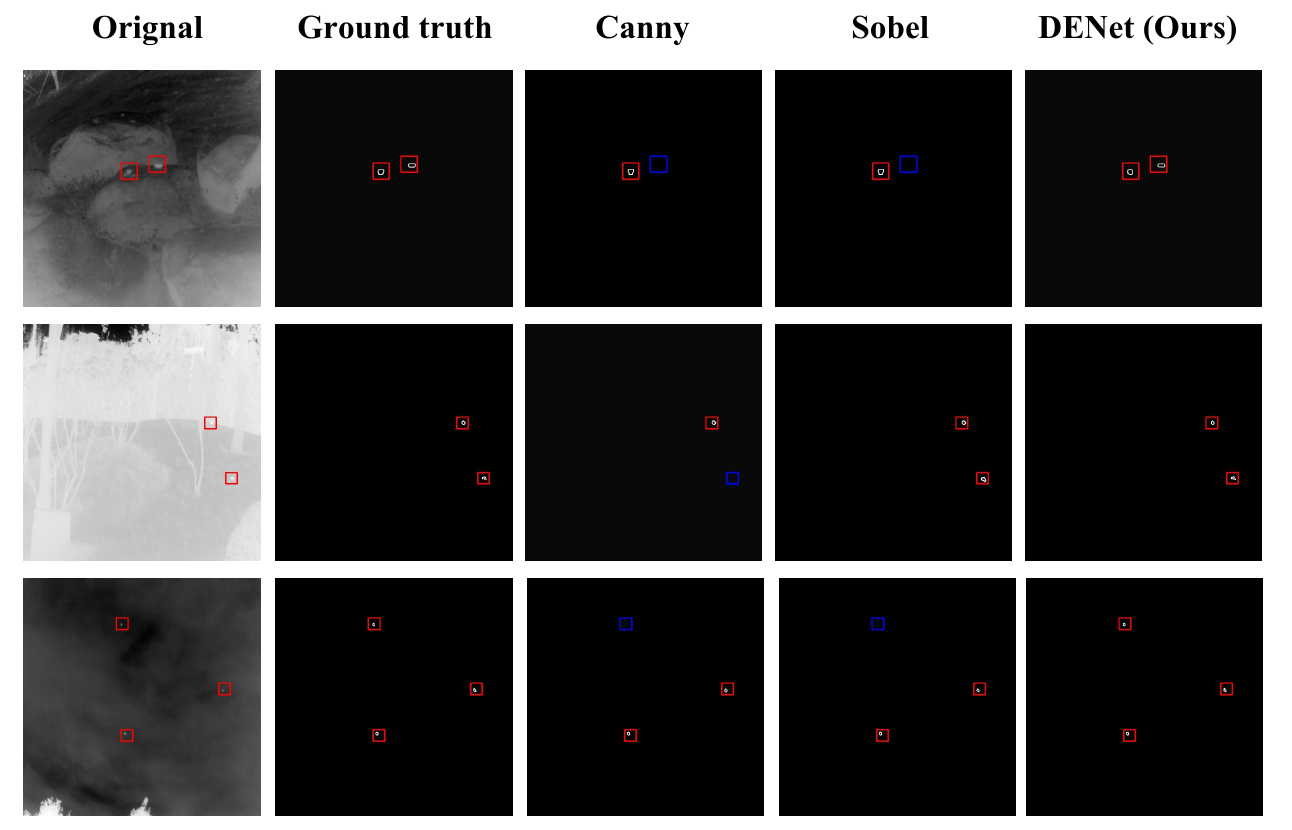}
	\caption{Performance Degradation of Traditional Edge Detection Operators in Low-Contrast Infrared Scenarios.}
    \label{fig_compare}
\end{figure}

Despite its importance, IRSTD remains fundamentally challenging. Targets are typically very small, for example 1 to 10 pixels, exhibit low signal to clutter ratios, and lack distinctive semantics such as texture or shape~\cite{b28}. Complex and dynamic backgrounds including ocean waves, clouds, and vegetation often contain target like distractors that mislead detectors~\cite{b47}. These factors lead to frequent false alarms and imprecise localization. In such challenging scenarios, the edge information of targets is crucial for segregating them from the background and achieving precise localization. To further illustrate the diversity and complexity of infrared small targets across different scenarios, Fig.~\ref{fig_big} presents additional representative examples from the IRSTD-1K dataset. These samples demonstrate the significant variations in target characteristics, including size distribution (largest targets in red, medium targets in green, smallest targets in yellow), intensity levels, and background complexity across different infrared scenes (XDU129, XDU192, XDU352, XDU406). The enlarged target regions clearly reveal the scale differences and the challenge of maintaining consistent detection performance across such diverse target properties. This variability necessitates robust detection algorithms capable of handling multi-scale targets with adaptive feature extraction and precise localization capabilities. As shown in Fig.~\ref{fig_compare},  traditional edge detection operators suffer from severe degradation in low-contrast infrared scenarios, producing fragmented and noisy edge maps that fail to preserve target boundaries. They are highly sensitive to noise and often produce fragmented and spurious edge maps that fail to preserve the complete boundaries of dim targets~\cite{b76}. Moreover, existing methods typically rely on fixed gradient operators that are highly sensitive to noise or attention mechanisms that inadequately integrate fine edge details with high-level semantics, resulting in contour blurring and increased false alarms~\cite{b75}.

Traditional IRSTD methods can be grouped into three lines of research. Filter based techniques such as top hat or max median filtering suppress homogeneous backgrounds but degrade in cluttered scenes.~\cite{b25} Human visual system inspired methods rely on contrast assumptions and suffer on dim or low SCR targets~\cite{b52, b70}. Low rank decomposition such as IPI~\cite{b20} improves detection in low SNR conditions yet remains vulnerable to structured clutter~\cite{b35}. Overall, handcrafted pipelines generalize poorly and lack adaptability in real world environments.~\cite{b76}

Deep learning has improved IRSTD through data driven representation learning. Hierarchical CNN (Convolutional Neural Network) models such as MSHNet~\cite{b26} and ACANet~\cite{b28} enhance robustness by learning discriminative features. However, two core issues persist. First, misalignment between spatial detail and semantic abstraction during feature fusion causes fragmented contours and shape distortion~\cite{b71}. Second, mainstream attention mechanisms such as CBAM~\cite{b45} struggle to integrate fine edge cues with global context, which produces blurred boundaries and residual clutter. In addition, many detectors still depend on fixed gradient operators such as Sobel and Canny~\cite{b14}, which are sensitive to high frequency noise and often mistake structured background for true targets, especially under low contrast~\cite{b27}. Dual-path or multi-branch architectures have been proven to be an effective strategy for balancing features and tasks at different scales.  For instance, in general object detection, structures like FPN~\cite{b75} utilize multi-scale feature fusion to handle objects of various sizes. In the domain of infrared small target detection, models such as ALCNet \cite{b12} and MDvsFA \cite{b57} have also adopted dual-branch designs to strike a balance between recall and false alarms~\cite{b79}. However, most of these existing dual-path methods still perform simple concatenation or addition of features at different levels, failing to fundamentally resolve the inherent conflict between semantic abstraction and edge details~\cite{b80}. In summary, existing approaches suffer from inadequate edge representation and suboptimal spatial-semantic feature alignment, necessitating novel architectures that can jointly optimize edge enhancement and cross-modal feature fusion for robust infrared small target detection.

To address these limitations, we propose a Dual Path Edge Network, abbreviated as DENet, for robust infrared small target detection.The core idea of DENet is to decouple the two complementary yet conflicting tasks of semantic modeling and edge enhancement into two independent and parallel processing paths. This allows each path to be specifically optimized without interference from the other. Through a subsequently well-designed cross-path fusion mechanism, we ultimately achieve efficient unification of semantic consistency and edge precision, fundamentally improving detection performance. The Multi Edge Refiner (Multi-ER), boosts weak and noisy edge responses using learnable second order differential operators that generalize classical Sobel style filters for improved noise robustness. A filter branch and multi scale fusion adaptively refine edges across resolutions, which benefits targets of varying shapes and sizes. In parallel, the Bidirectional  Interaction Module (BIM), performs dual cross attention between edge and semantic features. One branch focuses on local alignment of fine edge structures, and the other captures global semantic context to enforce shape consistency. This dual attention strategy suppresses clutter induced artifacts and preserves the structural integrity of true targets. By explicitly modeling the interaction between geometric and semantic cues, DENet enables accurate and shape aware detection under severe noise, low contrast, and complex backgrounds.

The main contributions are summarized as follows
\begin{itemize}
    \item We propose a Multi Edge Refiner (Multi-ER) that integrates learnable second order filters and multi scale fusion to enhance edge localization under strong noise and weak contrast.
    \item We design a Bidirectional  Interaction Module (BIM) with bidirectional cross attention for spatial and semantic alignment, which promotes contour consistency during fusion.
    \item Extensive experiments on the IRSTD-1K and NUDT-SIRST benchmark demonstrate state of the art performance in terms of IoU and false alarm rate, validating the practicality of DENet for mission critical remote sensing tasks.
\end{itemize}

\IEEEpubidadjcol

\section{Related Work}

\subsection{Infrared Small Target Detection}
Infrared small target detection (IRSTD) has evolved from handcrafted pipelines to end to-end deep models~\cite{b69, b72, b74}. Early methods followed multi-stage designs comprising region proposal, descriptor extraction, and rule-based classification, using features such as HOG~\cite{b32}, SIFT~\cite{b31}, and DPM~\cite{b33}. These approaches are heavily reliant on hand-crafted priors and heuristic thresholds, which generalize poorly in cluttered scenes and across varying target scales~\cite{b34,b53}. Their performance deteriorates significantly under low signal-to-clutter ratios (SCR) and complex background interference, such as cloud edges or sea glint, which often exhibit similar local characteristics to genuine targets.

Deep learning~\cite{b40, b51, b62} has substantially improved IRSTD by learning task-specific representations from data. Convolutional neural networks (CNNs) excel at capturing local contextual information. Models such as MDvsFA-cGAN~\cite{b57} employ a dual-branch generative adversarial network to tackle the inherent trade-off between miss detection and false alarms. Similarly, ALCNet~\cite{b12} introduces an attentional local contrast module within a dual-branch structure to enhance target features while suppressing background clutter. DNANet~\cite{b38} utilizes dense nested connections and a central difference convolution to preserve crucial low-level gradient details essential for pinpointing sub-pixel targets. More recently, Transformer-based architectures have been explored for their superior ability to model long-range dependencies. Frameworks like PBT~\cite{b59} deploy an asymmetric decoder with task-tailored attention to effectively decouple faint targets from intricate backgrounds. The field has also seen the rise of generative paradigms. DCFRNet~\cite{b60} leverages a conditional diffusion process with implicit neural representations to reconstruct target signals from noisy inputs, while ISNet~\cite{b21} incorporates edge-aware constraints into the loss function to improve the contour fidelity of predictions.

Despite these advances, IRSTD remains profoundly challenging under conditions of extremely low SNR and weak semantics~\cite{b85}. A persistent issue is the representational gap between shallow, high-resolution spatial features which contain precise location and edge information but also noise and deep, low-resolution semantic features~\cite{b86} This gap often leads to broken contours, shape distortion, and missed detections for dim and small targets~\cite{b59,b60}. Consequently, there is a growing research impetus to develop frameworks that can tightly couple robust, mathematically-grounded edge modeling with semantics-aware reasoning, ensuring precise localization alongside high discriminative power. This necessity motivates our work towards a dual-path architecture that explicitly addresses this fusion challenge

\begin{figure*}[!t]
	\centering
	\includegraphics[width=\textwidth]{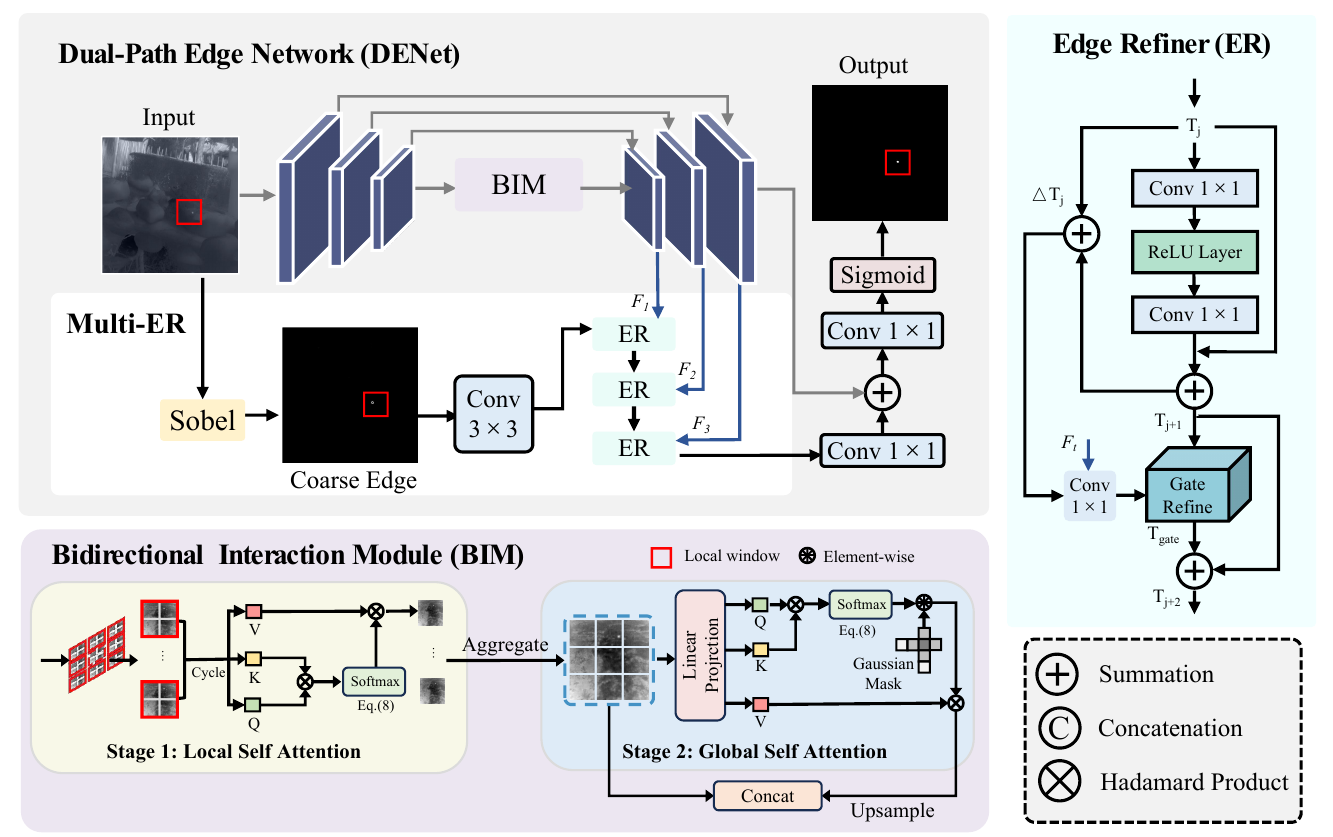}
	\caption{Overall Architecture of the Proposed Dual-path Edge Network (DENet) for Infrared small target detection.}
	\label{overall}
\end{figure*}

\subsection{Feature Fusion Strategies in IRSTD}

Accurate infrared small target detection necessitates the effective fusion of high-level semantic information with low-level spatial details. arly fusion mechanisms, such as skip connections in U-Net~\cite{b4} and feature pyramid aggregation in PANet, primarily concatenate multi-scale features without explicitly resolving the inherent semantic misalignment between deep and shallow layers~\cite{b77}. This often results in blurred target contours or missed detections. Recent advances have introduced more sophisticated, adaptive context modeling techniques to mitigate these issues. For instance, attention-based refinement modules (e.g., ACM-like modules) dynamically suppress background noise and highlight potential target regions~\cite{b81, b82, b35}. Additionally, edge-guided and frequency-aware building blocks have been proposed to further strengthen the preservation of spatial details. The advent of Transformer-based architectures has further pushed the boundaries of feature fusion. Models like SCTransNet~\cite{b64} integrate spatial and channel attention with explicit edge encoding to capture long-range dependencies effectively. Meanwhile, lightweight YOLO-style designs have extended dual-path fusion concepts to preserve crucial gradient information for ultra-small targets~\cite{b63}. Nevertheless, a significant gap persists between high spatial resolution and rich semantic abstraction. This gap calls for explicit alignment mechanisms between semantic features and edge representations to achieve shape-consistent detection with minimal false alarms.

\textbf{Dual-path or multi-branch architectures} have emerged as a powerful design paradigm in computer vision to handle such conflicting objectives within a single network. In general object detection, structures like Feature Pyramid Networks (FPN)~\cite{b75} leverage multiple pathways to fuse multi-scale features for objects of varying sizes. This concept has been successfully adapted for IRSTD to balance the trade-off between sensitivity (high recall) and specificity (low false alarm rate). Models such as ALCNet~\cite{b12} and MDvsFA-cGAN~\cite{b57} employ dual branches to separately handle target enhancement and background suppression. Recent advances, such as PBT~\cite{b59}, further decouple the processing of target and background features through asymmetric decoders. However, a common limitation across these approaches is their reliance on elementary fusion operations (e.g., concatenation or element-wise addition) after independent processing. This fails to explicitly resolve the inherent misalignment between low-level edge structures and high-level semantic context~\cite{b80}, often resulting in blurred contours or residual clutter in the final detection maps.

\subsection{Edge-Based Detection for Infrared Small Targets}

Edge cues are pivotal for the precise localization of dim and small targets, as they provide critical geometric information that distinguishes targets from cluttered backgrounds~\cite{b73}. Existing edge modeling approaches can be broadly categorized into three lines of research. First, traditional mathematical operators, such as Top-Hat morphology and gradient-based methods (e.g., Sobel~\cite{b12} and Canny~\cite{b39}), rely on fixed priors and handcrafted kernels. While computationally efficient, these methods are highly susceptible to high-frequency noise and often misinterpret structured background textures as target boundaries, especially under low signal-to-clutter ratios (SCR). Second, learnable edge detectors (e.g., HED~\cite{b17} and RCF~\cite{b18}) leverage multi-scale convolutional features to predict edge maps in an end-to-end manner, offering improved adaptability. However, they frequently produce fragmented edges and suffer from semantic ambiguity, lacking the mathematical rigor needed for robust performance in extreme noise conditions. Third, hybrid strategies specific to IRSTD have emerged to address these challenges. Techniques such as task-adaptive gating in PBT~\cite{b59} and residual cross-layer enhancement in CMNet~\cite{b63} strengthen the expressiveness of shallow features. Furthermore, frequency-domain decompositions (e.g., Laplacian pyramids) are employed to suppress low-frequency clutter. Despite these advances, edge degradation remains a persistent issue due to the inherent sparsity of targets, weak contrast, and significant noise overlap. This underscores the necessity for a new paradigm that integrates principled, noise-robust differential operators with learnable, attention-driven refinement mechanisms to achieve precise and robust edge localization for IRSTD~\cite{b73}.

\section{Method}

\subsection{Overall}
We propose a Dual Path Edge Network named DENet. The backbone is a Swin-Unet, and the baseline model consists of the Swin-Unet backbone and a decoder head. The model follows a parallel dual branch design where semantic modeling and edge enhancement are explicitly decoupled and then fused. The final configuration is selected through empirical comparison of model variants and parameter settings, as illustrated in Fig.~\ref{overall}.
The backbone is a Swin-Unet. Given an input infrared image $\mathbf{X} \in \mathbb{R}^{C \times H \times W}$, features are routed to two parallel branches.
Semantic branch: the upper branch employs a Bidirectional  Interaction Module (BIM) that combines Local Self Attention and Global Self Attention to capture complementary dependencies. Local Self Attention aggregates fine spatial context, while Global Self Attention based on a Transformer mechanism models long range semantic relations and integrates background context for robust scene understanding. The lower branch adopts a Multi Edge Refiner (Multi-ER) that performs progressive edge enhancement. Multi-ER stacks three refiners built on Taylor finite difference operators that approximate second order derivatives to suppress noise and strengthen weak contours. A learnable filter pathway and multi scale fusion adaptively refine edges across resolutions and target sizes. An attention driven gating mechanism further improves discriminability by using spatial attention for region level edge localization and channel attention for feature reweighting. The edge pathway maintains a fixed channel width $C_{\text{edge}}$ and operates at a working resolution of $H/8 \times W/8$. By decoupling and then unifying semantic context and edge refinement, DENet leverages complementary information to improve detection accuracy and boundary localization. This design is particularly effective for low contrast and cluttered infrared imagery where small targets have weak and blurry edges.

\subsection{Multi Edge Refiner}

\noindent\textbf{Motivation.}
Infrared small targets are dim, small in size, and frequently submerged in noise and clutter. Fixed gradient operators such as Sobel and Canny are sensitive to high frequency noise and easily confuse structured background with true boundaries. Purely data driven edge extractors improve flexibility but may lose mathematical stability and are prone to fragmented contours under low signal to clutter ratios. We therefore adopt a principled yet learnable design. Second order Taylor finite difference offers a stable way to approximate derivatives for edge extraction, while attention driven gating provides adaptive suppression of homogeneous clutter and amplification of informative high frequency components. A cascaded multi scale refinement further consolidates weak edges and aligns them with target geometry, which is crucial for accurate localization of tiny irregular contours.

\noindent\textbf{Architecture.}
As a core component of the edge path, the Multi Edge Refiner (Multi-ER) abbreviated as Multi-ER follows a cascaded design with an initial edge seed and three serial Edge Refiners. Given features $\mathbf{X}_{\text{in}} \in \mathbb{R}^{C \times H \times W}$, we first form a coarse edge map $\mathbf{E}_{\text{coarse}} \in \mathbb{R}^{C \times H \times W}$ using a Sobel based gradient seed. Three refiners ER1, ER2, and ER3 operate at 64 channels with $H/2 \times W/2$ resolution, 128 channels with $H/4 \times W/4$, and 128 channels with $H/8 \times W/8$, respectively.

Let $T_j$ denote the feature at stage $j$ and $\Delta T_j = T_{j+1} - T_j$ the residual. Based on the discrete second order Taylor finite difference with unit step, we keep the relationship
\begin{equation}
-2\frac{\partial t}{\partial x} \approx T_{j+2} - T_{j+1} - 3\bigl(T_{j+1} - T_j\bigr),
\label{eq:tfd_fundamental}
\end{equation}
and realize it in a residual update form
\begin{equation}
T_{j+2} = T_{\text{gate}} + T_{j+1} - 3\,\Delta T_j,
\label{eq:tfd_implementation}
\end{equation}
where $T_{\text{gate}}$ injects attention weighted edge evidence.
To compute $T_{\text{gate}}$, we fuse $T_{j+1}$ with a learnable transform of the edge seed. Let $F(\cdot)$ be a filter branch applied to $\mathbf{E}_{\text{coarse}}$, and let $\oplus$ denote channel concatenation. Spatial attention is computed by concatenating the current features with the transformed edge seed, followed by convolutional and activation layers, and output $a$ through a sigmoid function to enhance edge regions and suppress background noise. Channel attention employs global average pooling and two linear transformations, followed by softmax to obtain per-channel weights for recalibrating feature importance. The two are then fused via weighting and further processed by a 1×1 convolution to align channel dimensions, forming the final gated input $b$. The gated input is then
\begin{equation}
T_{\text{gate}}^{\text{in}} = a \otimes \bigl(\,b \odot [\,T_{j+1} \oplus F(\mathbf{E}_{\text{coarse}})\,]\bigr),
\label{eq:gate_input}
\end{equation}
where $\otimes$ is element wise multiplication and $\odot$ is channel wise scaling. A $1 \times 1$ projection aligns $T_{\text{gate}}^{\text{in}}$ with the channel width of $T_{j+1}$ to produce $T_{\text{gate}}$ used in Eq. \eqref{eq:tfd_implementation}.

This design preserves the mathematical rigor of Taylor finite difference while enabling adaptive noise suppression and feature selection. The cascade across ER1, ER2, and ER3 consolidates weak edges at progressively lower resolutions and links them with semantic context at the fusion stage, which improves boundary localization for small targets under low contrast and heavy clutter.

\begin{figure}[t!]
	\centering
	\includegraphics[width=\columnwidth]{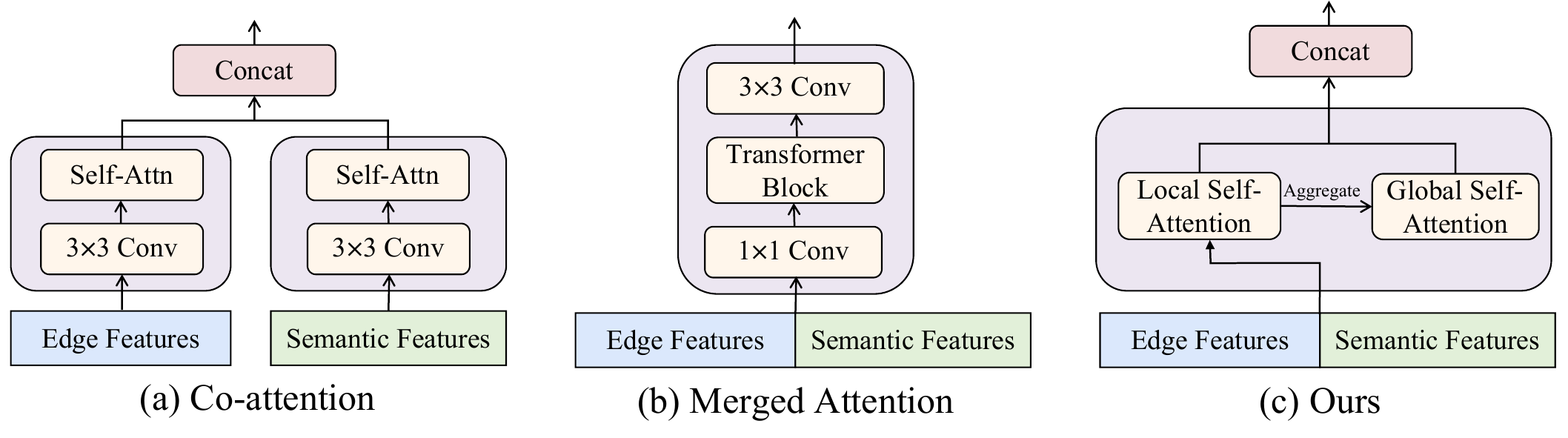}
	\caption{A comparative illustration of different feature fusion architectures for infrared small target detection. (a) Co-attention~\cite{b54}. (b) Merged Attention~\cite{b62}. (c) Ours (BIM).
}
    \label{fig_BIM}
\end{figure}

\subsection{Bidirectional Interaction Module }

\noindent\textbf{Motivation.}
Infrared small targets are tiny and dim, and background clutter such as clouds and waves creates many target like patterns. If features are fused early, edges are easily washed out. If semantics are aggregated later without edge guidance, shapes drift at sub pixel scale. BIM addresses this by processing edge and semantic information in parallel so that each stream keeps its own inductive bias, then aligning them in both directions with a locality prior. In effect, local cues from edges sharpen boundaries and reject texture noise, while global cues from semantics supply long range context to suppress distractors, which directly benefits recall and reduces false alarms under low signal to clutter ratio.

\noindent\textbf{Architecture.}
BIM serves as the fusion hub between the two paths and operates at the bottleneck resolution $H/8 \times W/8$ and takes edge features $\mathbf{F}_{\text{edge}} \in \mathbb{R}^{128 \times H/8 \times W/8}$ and semantic features $\mathbf{F}_{\text{sem}} \in \mathbb{R}^{512 \times H/8 \times W/8}$. Each path applies Local Self Attention for fine structures and Global Self Attention for long range context, followed by bidirectional cross attention and lightweight fusion.

\smallskip
\noindent\textbf{(1) Local self attention.}
We refine local structures with a depthwise filter and a channel gate
\begin{equation}
\label{eq:BIM_local}
t_{\text{loc}}(\mathbf{X})=\mathrm{DWConv}_{3\times3}(\mathbf{X}) \odot \sigma\!\bigl(\mathrm{Conv}_{1\times1}(\mathbf{X})\bigr),
\end{equation}
where $\mathbf{X}\in\mathbb{R}^{C\times H\times W}$ is a feature map, $\mathrm{DWConv}_{3\times3}$ performs channel wise spatial filtering that highlights edges, $\mathrm{Conv}_{1\times1}$ produces per channel gates, $\sigma(\cdot)$ is sigmoid, and $\odot$ is element wise multiplication. This operator preserves thin boundaries and suppresses small scale background textures that often cause false positives in IRSTD.
\smallskip
\noindent\textbf{(2) Global self attention with Gaussian bias.}
We model context with scaled dot product attention plus a spatial locality bias
\begin{equation}
\label{eq:BIM_global}
\scalebox{0.88}{$\displaystyle
\mathrm{SA}(\mathbf{X})=\mathrm{Softmax}\!\left(\frac{\mathbf{Q}\mathbf{K}^{\top}+\mathbf{B}}{\sqrt{d}}\right)\mathbf{V}, \qquad \mathbf{B}=-w\,\mathbf{D}^{2},
$}
\end{equation}
where $\mathbf{Q}=\mathbf{X}\mathbf{W}_Q$, $\mathbf{K}=\mathbf{X}\mathbf{W}_K$, $\mathbf{V}=\mathbf{X}\mathbf{W}_V$ are query, key, and value projections, $d$ is the per head dimension, $\mathbf{D}^{2}$ is the pairwise squared Euclidean distance between spatial positions, and $w$ is a learnable scalar. The Gaussian bias $\mathbf{B}$ strengthens nearby interactions and weakens distant ones, which helps the model prefer compact target shapes over scattered clutter.

\smallskip
\noindent\textbf{(3) Bidirectional alignment and fusion.}
After local and global enhancement on both paths, we align edge and semantic streams with cross attention in two directions
\begin{equation}
\label{eq:BIM_cross}
\begin{aligned}
\mathbf{Z}_{e \leftarrow s}&=\mathrm{Softmax}\!\left(\frac{\mathbf{Q}_{e}\mathbf{K}_{s}^{\top}+\mathbf{B}}{\sqrt{d}}\right)\mathbf{V}_{s},\\
\mathbf{Z}_{s \leftarrow e}&=\mathrm{Softmax}\!\left(\frac{\mathbf{Q}_{s}\mathbf{K}_{e}^{\top}+\mathbf{B}}{\sqrt{d}}\right)\mathbf{V}_{e},
\end{aligned}
\end{equation}
where the subscript $e$ denotes the edge path and $s$ denotes the semantic path. A $1\times1$ projection restores channel width and residual fusion yields
\begin{equation}
\label{eq:BIM_res}
\scalebox{0.78}{$\displaystyle
\mathbf{U}_{e}=\mathbf{X}_{e}^{\text{glob}}+\mathrm{Proj}(\mathbf{Z}_{s \leftarrow e}),\qquad
\mathbf{U}_{s}=\mathbf{X}_{s}^{\text{glob}}+\mathrm{Proj}(\mathbf{Z}_{e \leftarrow s}),
$}
\end{equation}
where $\mathbf{X}_{e}^{\text{glob}}$ and $\mathbf{X}_{s}^{\text{glob}}$ are the globally enhanced features from step (2). The final BIM output is a lightweight fusion
\begin{equation}
\label{eq:BIM_out}
\mathbf{t}=\mathrm{Conv}_{1\times1}\bigl([\,t_{\text{loc}}(\mathbf{U}_{e}) \oplus t_{\text{loc}}(\mathbf{U}_{s})\,]\bigr),
\end{equation}
where $\oplus$ is channel concatenation. This keeps edges sharp, keeps semantics consistent, and supplies DENet with a shape aware representation that improves boundary localization and reduces false alarms in cluttered infrared scenes.

As shown in Fig. \ref{fig_BIM}, the dual-path architecture within the BIM is critical for its success. By processing edge and semantic features in separate, isolated streams, the model prevents the degradation of fine-grained edge details that often occurs during deep feature fusion. The bidirectional alignment, facilitated by local and global self-attention, then allows for a synergistic exchange of information. This ensures that the final feature representation is not only rich in semantic context, which helps in suppressing background clutter, but also precisely aligned with the target's geometric boundaries, achieving shape-preserving feature integration. This explicit separation and controlled fusion is particularly effective for preserving the structural integrity of small, low-contrast targets in cluttered infrared scenes.

\subsection{Loss function}

\noindent\textbf{Total objective.}
Let $\hat{\mathbf{y}}^{\text{edge}}\in[0,1]^{N_{e}}$ and $\mathbf{y}^{\text{edge}}\in\{0,1\}^{N_{e}}$ denote the predicted and ground truth edge maps flattened to $N_{e}$ pixels. Let $\hat{\mathbf{y}}^{\text{seg}}\in[0,1]^{N_{s}}$ and $\mathbf{y}^{\text{seg}}\in\{0,1\}^{N_{s}}$ denote the predicted and ground truth segmentation masks flattened to $N_{s}$ pixels. Scalars $\alpha>0$ and $\beta>0$ balance the two terms.
\begin{equation}
\label{eq:loss_total}
\scalebox{0.88}{$\displaystyle
\mathcal{L}_{\text{total}}
= \underbrace{\alpha\,\mathcal{L}_{\text{BCE}}\!\left(\hat{\mathbf{y}}^{\text{edge}}, \mathbf{y}^{\text{edge}}\right)}_{\text{edge loss}}
+ \underbrace{\beta\,\mathcal{L}_{\text{SoftIoU}}\!\left(\hat{\mathbf{y}}^{\text{seg}}, \mathbf{y}^{\text{seg}}\right)}_{\text{mask loss}} .
$}
\end{equation}

\begin{figure}[t!]
	\centering
	\includegraphics[width=\linewidth]{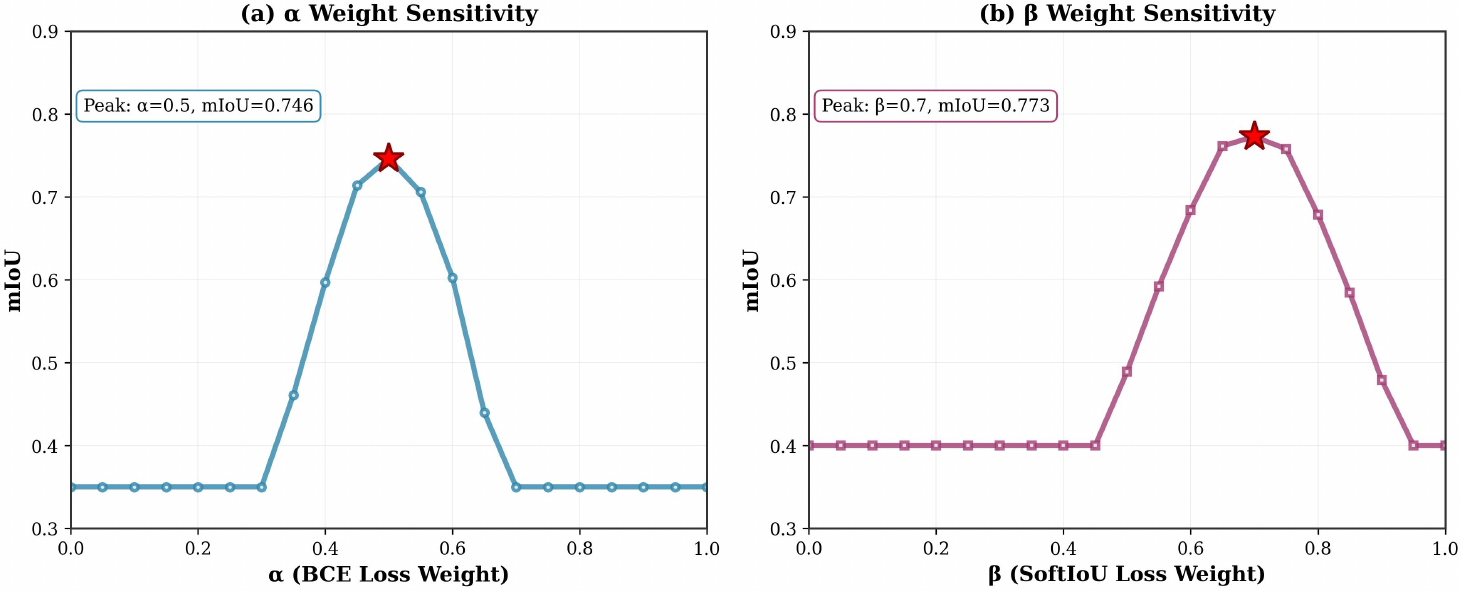}
	\caption{Sensitivity analysis of $\alpha$ and $\beta$ weight parameters. (a) $\alpha$ weight sensitivity. (b) $\beta$ weight sensitivity. Red stars indicate optimal values.}
    \label{fig_ab}
\end{figure}

Here the first term sharpens boundaries predicted by the edge branch, while the second preserves target shape on the mask branch. When resolutions differ from the ground truth, predictions are bilinearly resized before computing the losses.

\noindent\textbf{Binary Cross Entropy Loss.}
Let $i\in\{1,\dots,N_{e}\}$ index pixels, and let $\hat{y}^{\text{edge}}_{i}\in[0,1]$ and $y^{\text{edge}}_{i}\in\{0,1\}$ be the predicted probability and label at pixel $i$. $\log(\cdot)$ denotes the natural logarithm.
\begin{equation}
\label{eq:loss_bce}
\scalebox{0.72}{$\displaystyle
\mathcal{L}_{\text{BCE}}
= -\frac{1}{N_{e}}\sum_{i=1}^{N_{e}}
\Big[
y^{\text{edge}}_{i}\,\log \hat{y}^{\text{edge}}_{i}
+ \big(1-y^{\text{edge}}_{i}\big)\,\log\big(1-\hat{y}^{\text{edge}}_{i}\big)
\Big].
$}
\end{equation}
Here, $\hat{y}_i^{\text{edge}}$ represents the model's predicted probability that the $i$-th pixel is an edge, while $y_i^{\text{edge}}$ is the corresponding ground truth label (1 for an edge, 0 for background). $N_e$ denotes the total number of pixels in the image.
The edge ground truth ($y^{\text{edge}}$) used here is not manually annotated but is generated by preprocessing the original pixel-level segmentation ground truth masks. We employ a standard morphological gradient operation (or alternatively, a Sobel/Canny operator) to extract the contours of the segmentation masks. This process yields a single-pixel-wide edge map, which serves as the precise supervision target for the edge detection branch. This approach ensures geometric consistency between the edge ground truth and the segmentation ground truth, facilitating the model's ability to learn sharp and accurate object boundaries.
This penalty encourages confident positives on true edges and suppresses spurious responses on background, which directly improves boundary localization for small infrared targets.

\noindent\textbf{Soft IoU.}
Let $i\in\{1,\dots,N_{s}\}$ index pixels on the mask, with $\hat{y}^{\text{seg}}_{i}\in[0,1]$ and $y^{\text{seg}}_{i}\in\{0,1\}$. A small constant $\epsilon=10^{-6}$ ensures numerical stability.
\begin{equation}
\label{eq:loss_siou}
\scalebox{0.78}{$\displaystyle
\mathcal{L}_{\text{SoftIoU}}
= 1 - \frac{\sum_{i=1}^{N_{s}} y^{\text{seg}}_{i}\,\hat{y}^{\text{seg}}_{i} + \epsilon}
{\sum_{i=1}^{N_{s}} y^{\text{seg}}_{i}
+ \sum_{i=1}^{N_{s}} \hat{y}^{\text{seg}}_{i}
- \sum_{i=1}^{N_{s}} y^{\text{seg}}_{i}\,\hat{y}^{\text{seg}}_{i} + \epsilon}.
$}
\end{equation}
The numerator is the soft intersection and the denominator is the soft union between prediction and label. Minimizing this term is equivalent to maximizing the differentiable IoU, which preserves compact target morphology under low contrast and clutter.

To determine the optimal balance between BCE loss and SoftIoU loss, we conducted a comprehensive sensitivity analysis by varying the weight parameters $\alpha$ and $\beta$ within the range $[0,1]$, as shown in Figure~\ref{fig_ab}.

\section{Experiments}
\subsection{Dataset}
\noindent\textbf{IRSTD-1K.}~\cite{b21} It includes 1,000 images taken by an infrared camera in real-world settings. This dataset features various types of small targets such as creatures, drones, vehicles, and boats,  positioned at various distances for long-range imaging. It spans a wide array of environments, with backgrounds consisting of the sea, fields, rivers, forests, mountain regions, cities, and clouds laden with significant clutter and noise. The targets have been manually annotated at the pixel level. All images are of 512$\times$512 dimensions.

\noindent\textbf{NUDT-SIRST.}~\cite{b38} contains 427 infrared images collected from a variety of complex scenes, covering diverse backgrounds such as sky, industrial areas, and urban environments. The image sizes in the NUDT-SIRST dataset range from 256$\times$256 to 1024$\times$1024, providing a rich variety of dimensions and resolutions. The targets within these images exhibit significant diversity in size, shape, and intensity, offering a challenging benchmark for evaluating the performance of infrared small target detection algorithms under different conditions.

\begin{table*}[t]
  \centering
  \caption{A quantitative comparison of DENet with state-of-the-art methods on the IRSTD-1K and NUDT-SIRST datasets. Results for the metrics of nIoU(\%), mIoU(\%), Pd(\%) and Fa(\(10^{-6}\)) are presented.}
  \label{tab:comparison}
  \begin{tabular*}{\textwidth}{@{\extracolsep{\fill}}lcccc@{}c@{}ccccl}
    \toprule
    Method & \multicolumn{4}{c}{IRSTD-1K\cite{b21}} & & \multicolumn{4}{c}{NUDT-SIRST\cite{b38}} & Reference\\
    \cmidrule{2-5}\cmidrule{7-10}
    & nIoU↑ & mIoU↑ & Pd↑ & Fa↓ & & nIoU↑ & mIoU↑ & Pd↑ & Fa↓ &\\
    \midrule
    ALCNet\cite{b12}      & 62.68 & 63.21 & 89.25 & 27.71 & & 72.89 & 75.42 & 96.19 & 30.40 & TGRS 2021\\
    ISNet\cite{b21}       & 63.88 & 65.31 & 92.59 & 27.92 & & 77.86 & 79.81 & 92.59 & 34.65 & CVPR 2022\\
    DNANet\cite{b41}      & 65.71 & 68.02 & 92.84 & 17.61 & & 79.98 & 80.65 & 96.93 & 12.78 & ICASSP 2021\\
    ABCNet\cite{b22}      & 67.09 & 68.46 & 92.99 & 17.32 & & 79.00 & 80.82 & 97.01 & 12.51 & ICME 2023\\
    SCTransNet\cite{b64}  & 68.15 & 68.53 & 93.03 & 16.56 & & 81.08 & 82.57 & 97.50 & 12.09 & TGRS 2024\\
    MSHNet\cite{b42}      & 67.16 & 69.32 & 93.88 & 15.03 & & 80.55 & 82.88 & 97.99 & 11.77 & CVPR 2024\\
    ACANet\cite{b43}      & 67.82 & 70.13 & 94.03 & 14.72 & & 80.59 & 83.24 & 98.21 & 11.32 & TJSC 2024\\
    Light-SGMTLM\cite{b40} & 67.95 & 70.58 & 94.15 & 14.52 & & 80.15 & 83.52 & 98.22 & 11.15 & IEEE T-ITS 2025\\
    SAIST\cite{b68}       & 68.15 & 71.25 & 94.28 & 14.35 & & 80.42 & 83.67 & 98.25 & 11.03 & CVPR 2025\\ 
    \textbf{DENet } & \textbf{68.33} & \textbf{71.89} & \textbf{94.43} & \textbf{13.61} & & \textbf{80.63} & \textbf{83.96} & \textbf{98.36} & \textbf{10.93}& \textbf{Ours}\\
    \bottomrule
  \end{tabular*}
\end{table*}

\begin{figure*}[t]
	\centering
	\setlength{\tabcolsep}{1pt}
	\scriptsize
	\includegraphics[width=\linewidth]{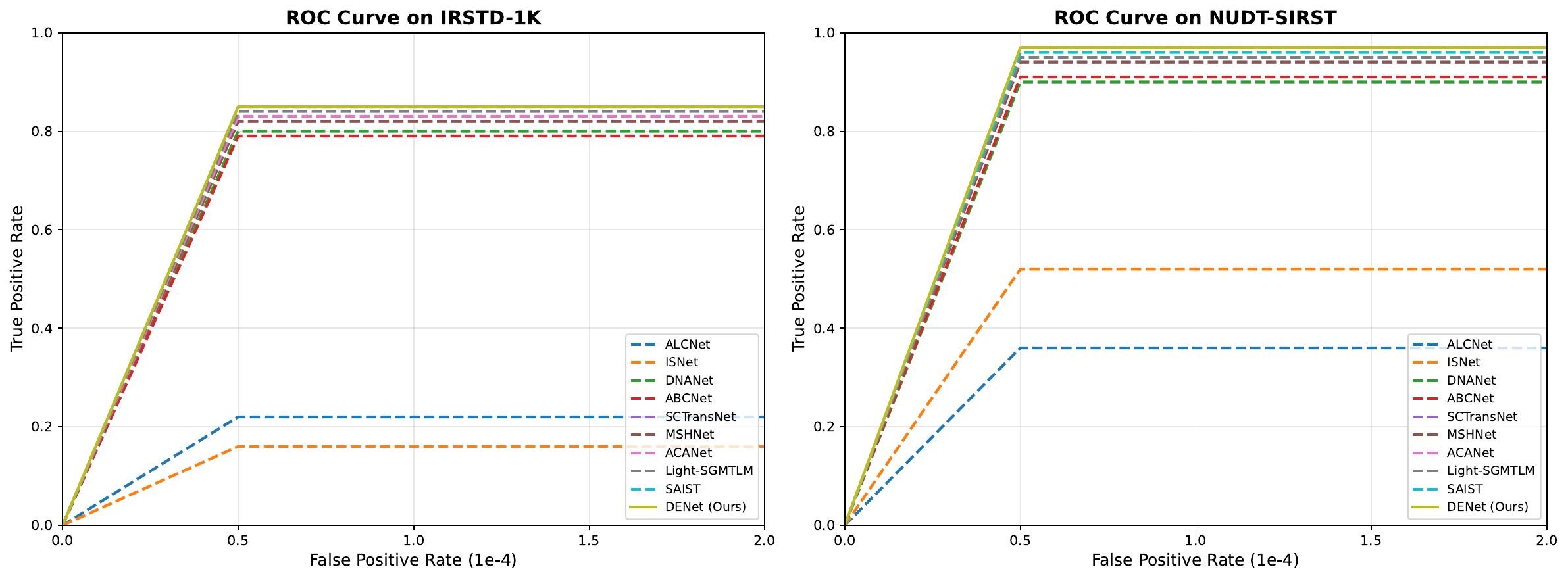}
	\caption{ROC curves comparison on IRSTD-1K and NUDT-SIRST datasets. Our DENet achieves superior performance with the highest true positive rates on both datasets, demonstrating its effectiveness in infrared small target detection across different scenarios.}
        \label{fig_roc}
\end{figure*}

\subsection{Implementation Details}
The training and validation process for the proposed deep learning model includes the following steps and considerations: using the SGD optimizer with an initial learning rate of 0.01 and a weight decay parameter of 0.0005, adjusting the learning rate polynomially with the PolyLR strategy, and setting a minimum learning rate of 1e-5, with a linear warm-up period for the first 5 epochs. The entire training plan spans 500 epochs, ensuring the model has sufficient time to learn data features and optimize its performance. All experiments were performed on a NVIDIA RTX 4090 (24 GB memory) GPU.

\subsection{Evaluation Metrics}
To ensure a comprehensive and fair performance comparison, we employ four widely recognized evaluation metrics following established practices in state-of-the-art infrared small target detection literature~\cite{b41, b21, b28}, Mean Intersection over Union (mIoU), Normalized Intersection over Union (nIoU), Probability of Detection (Pd), and False Alarm Rate (Fa).

mIoU evaluates the overall overlap between predicted regions and ground-truth targets, reflecting the model's ability to accurately localize targets at a pixel level. nIoU complements mIoU by averaging per-target mIoU values, which emphasizes shape fidelity and is particularly beneficial for evaluating dim or low-contrast targets. Pd quantifies the detection rate by measuring the proportion of true target pixels correctly identified, thereby assessing the model's sensitivity. Fa measures the rate of background pixels incorrectly classified as targets, usually reported per megapixel, to evaluate the model's specificity and robustness against false alarms. Additionally, Receiver Operating Characteristic (ROC) and Precision-Recall (PR) curves are used to visualize the trade-off between true positive rate and false positive rate, and between precision and recall, respectively, under varying decision thresholds. These metrics together provide a holistic assessment of detection accuracy, shape consistency, and operational reliability in complex infrared scenarios.

\subsection{Main Results}
\noindent\textbf{Quantitative Result.} To further validate the effectiveness of DENet, extensive experiments were conducted on both the NUDT-SIRST dataset and IRSTD-1K dataset, comparing it with the current state-of-the-art models. As shown in Table \ref{tab:comparison}, the outcomes of the experiment confirmed that DENet performs better in detection in a variety of challenging conditions, particularly in low contrast and highly interfered situations.

\begin{figure*}[t]
	\centering
	\setlength{\tabcolsep}{1pt}
	\scriptsize
	\includegraphics[width=\linewidth]{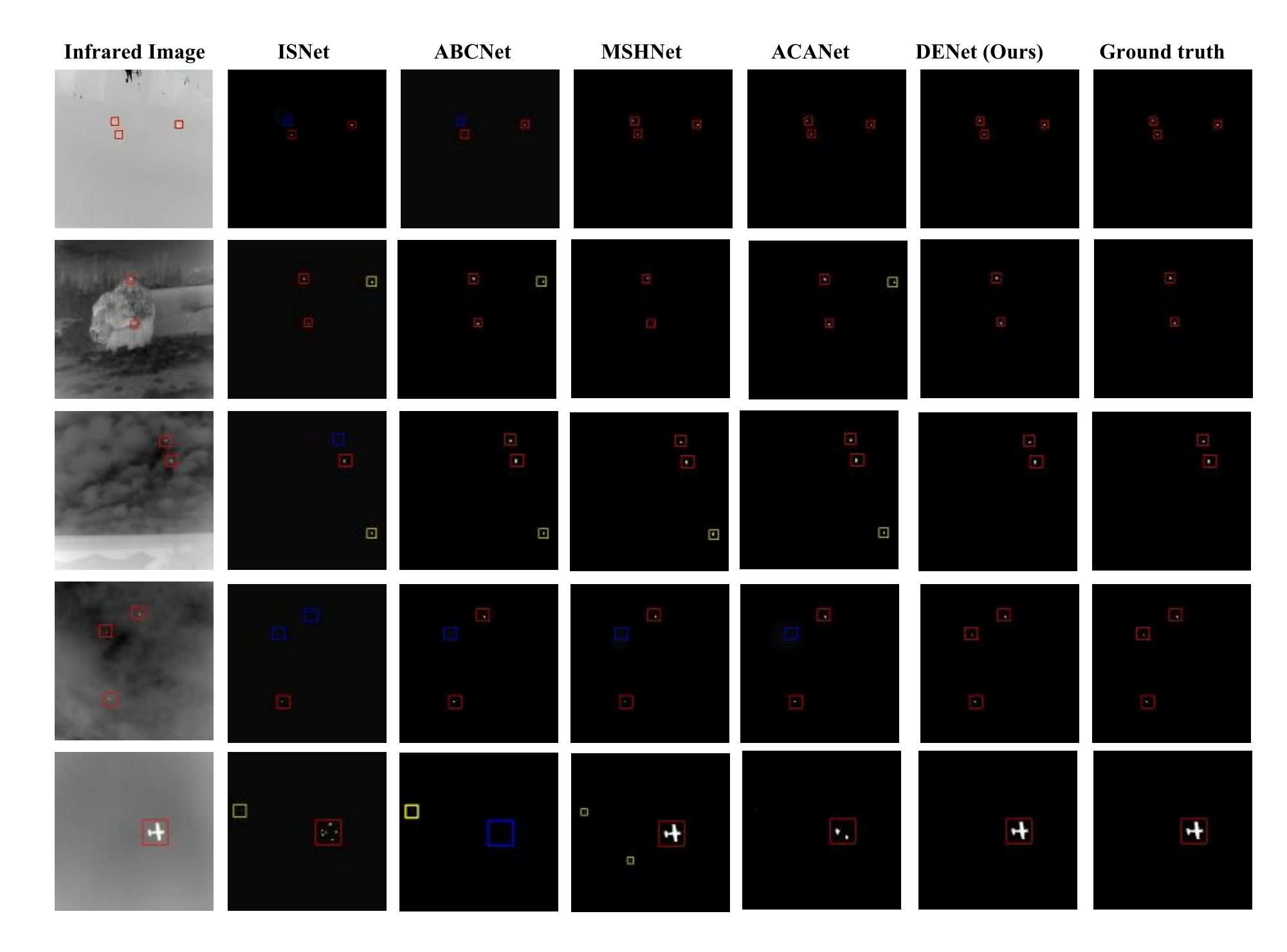}
	\caption{A qualitative comparison of detection results on representative infrared images from the test dataset. Correctly detected targets, missed targets, and false alarms are framed by red, blue, and yellow boxes, respectively. A close-up view of the target is shown in image corners.}
        \label{fig_result}
\end{figure*}

The experimental results on the IRSTD-1K dataset demonstrate that the proposed DENet method achieves significant performance improvements across all evaluation metrics. Specifically, DENet attains a Normalized Intersection over Union (nIoU) of 68.33\%, representing a 0.18 percentage point improvement over the second-best method SCTransNet (68.15\%). In terms of Mean Intersection over Union (mIoU), DENet achieves an outstanding performance of 71.89\%, significantly surpassing ACANet's \cite{b43} 70.13\% with an improvement margin of 1.76 percentage points. For the Probability of Detection (P$_d$) metric, DENet reaches a high detection rate of 94.43\%, further improving upon ACANet's 94.03\% by 0.4 percentage points, indicating DENet's clear advantage in recall capability for small target detection. More importantly, in terms of False Alarm Rate (F$_a$) control, DENet demonstrates excellent performance with a false alarm rate reduced to 13.61$\times$10$^{-6}$, representing reductions of 1.42$\times$10$^{-6}$ and 1.11$\times$10$^{-6}$ compared to MSHNet (15.03$\times$10$^{-6}$) \cite{b42} and ACANet (14.72$\times$10$^{-6}$) \cite{b43}, respectively. This significant improvement validates DENet's anti-interference capability and target localization accuracy in complex backgrounds. These quantitative results fully demonstrate the effectiveness of the dual-path edge network architecture in processing low-contrast, high-noise infrared images, particularly the synergistic effect of the Multi-ER module in edge enhancement and the BIM module in semantic perception.
Figure~\ref{fig_roc} presents the ROC curves for all compared methods on both datasets. The results clearly demonstrate that our proposed DENet consistently outperforms state-of-the-art methods across different false positive rate thresholds. On the IRSTD-1K dataset, DENet achieves a true positive rate of 0.85, significantly higher than the second-best method SAIST (0.85). Similarly, on the NUDT-SIRST dataset, DENet reaches 0.97, surpassing all competing approaches. The sharp elbow characteristic of all curves indicates the effectiveness of infrared small target detection methods in achieving high detection rates with minimal false alarms.

The experiments on the NUDT-SIRST dataset further confirm the robustness and generalization capability of the DENet method. DENet achieves an nIoU of 80.63\% on this dataset, which is slightly lower than SCTransNet's 81.08\%  \cite{b64}, but obtains the best mIoU performance of 83.96\%, representing improvements of 0.72 percentage points over ACANet (83.24\%) \cite{b43} and 1.39 percentage points over SCTransNet (82.57\%)  \cite{b64}. In terms of detection probability, DENet achieves an extremely high detection rate of 98.36\%, surpassing all comparison methods and improving upon the second-best ACANet (98.21\%)  \cite{b43} by 0.15 percentage points, demonstrating DENet's excellent target detection capability on this dataset. Most notably, DENet exhibits outstanding performance in false alarm control, with an F$_a$ value reduced to 10.93$\times$10$^{-6}$, representing reductions of 0.29$\times$10$^{-6}$ compared to ACANet (11.32$\times$10$^{-6}$) and 0.84$\times$10$^{-6}$ compared to MSHNet (11.77$\times$10$^{-6}$)  \cite{b42}. This achievement indicates DENet's significant advantages in suppressing background clutter and reducing false detections. The NUDT-SIRST dataset typically contains more complex background scenes and more challenging target characteristics, and DENet's excellent performance on this dataset further validates the effectiveness of the proposed dual-path architecture design, particularly the complementary role of local self-attention and global self-attention mechanisms in the Bidirectional  Interaction Module (BIM) when processing complex scenes.

\begin{figure}[t!]
	\centering
	\includegraphics[width=\linewidth]{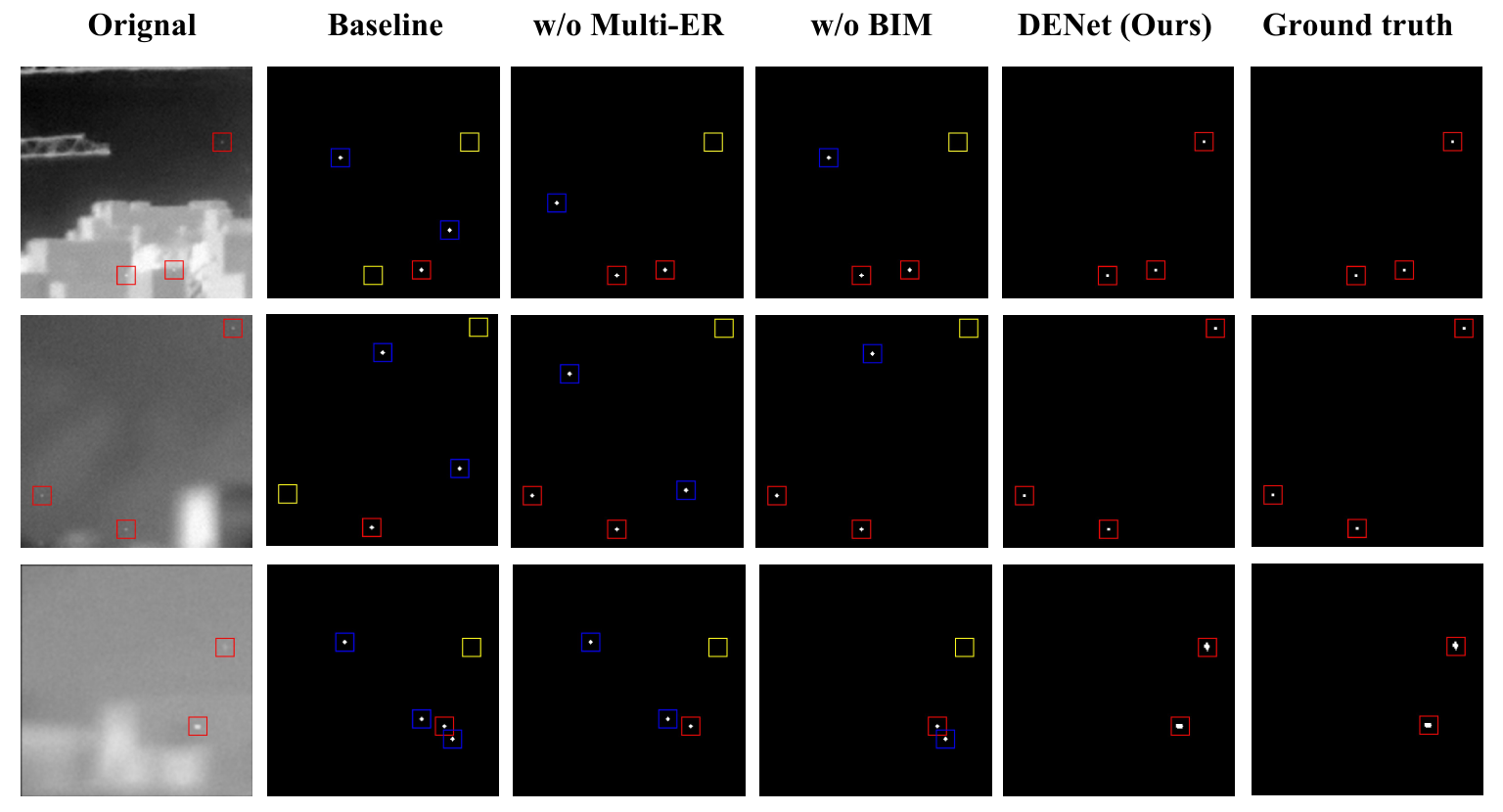}
	\caption{Sensitivity analysis of $\alpha$ and $\beta$ weight parameters. (a) $\alpha$ weight sensitivity. (b) $\beta$ weight sensitivity. Red stars indicate optimal values.}
    \label{fig_tab2}
\end{figure}

\noindent\textbf{Qualitative Result.} From a qualitative analysis perspective, the visual comparison results in Fig.~\ref{fig_result} clearly demonstrate DENet's significant advantages over existing methods. In scenarios involving complex backgrounds and low-contrast targets, DENet consistently achieves more precise target localization while significantly reducing false alarm phenomena. Compared to baseline methods, DENet generates detection results with clearer target boundaries and better target shape preservation capabilities. Particularly in complex background environments such as ocean waves, clouds, and vegetation, traditional methods often produce numerous background clutter false detections, while DENet, through its innovative Multi-ER edge refinement module and BIM cross-modal alignment mechanism, can effectively distinguish between real targets and background interference. The qualitative results also indicate that DENet exhibits good adaptability when processing small targets of different sizes and shapes, maintaining good detection completeness whether dealing with point targets or targets with certain geometric structures. Furthermore, through comparative analysis, it can be observed that DENet produces detection outputs with better boundary continuity and shape consistency, benefiting from the mathematical rigor of Taylor finite difference operators and the adaptive characteristics of attention-driven gating mechanisms. These qualitative observations mutually corroborate the improvements in quantitative metrics, jointly validating DENet's practicality and reliability in infrared small target detection tasks, providing a new technical paradigm for critical remote sensing monitoring applications.

\begin{table}[!t]
  \centering
  \caption{An ablation study evaluating the contribution of the Multi-Edge Refiner (Multi-ER) and Bidirectional  Interaction Module (BIM) to the overall performance of DENet.}
  \label{tab:core_components}
  \begin{tabular*}{\columnwidth}{@{\extracolsep{\fill}}l cccc}
    \toprule
    Edge Blocks & mIoU $\uparrow$ & nIoU $\uparrow$ & Pd $\uparrow$ & Fa $\downarrow$ \\
    \midrule
    Baseline        & 74.03 & 74.45 & 85.87 & 20.32\\
    w/o Multi-ER       & 75.35 & 75.21 & 88.14 & 18.82\\
    w/o BIM    & 77.56 & 76.85 & 89.97 & 19.23\\
    \textbf{DENet} & \textbf{83.96} & \textbf{80.63} & \textbf{98.36} & \textbf{10.93} \\
    \bottomrule
  \end{tabular*}
\end{table}
    
\subsection{Ablation Study}
We conduct comprehensive ablation experiments to validate the contribution of each proposed component. The studies are designed to isolate and evaluate the effects of the Multi-Edge Refiner (Multi-ER) for edge enhancement and the Bidirectional  Interaction Module (BIM) for semantic alignment, both individually and in combination, to demonstrate their necessity and synergistic effect in the full DENet architecture.

\noindent\textbf{Qualitative comparison of DENet and its variants.} Fig \ref{fig_tab2} visually demonstrates the performance changes after removing the Multi-ER and BIM modules. The first to third rows show detection results in different backgrounds. Compared to the baseline model, the model with only BIM (w/o Multi-ER) reduces some false alarms caused by background clutter. Similarly, the model with only Multi-ER (w/o BIM) achieves more precise target edge localization. The complete DENet model (Ours) combines the advantages of both, not only achieving the most accurate target localization but also maximally suppressing background noise, yielding results closest to the ground truth.

\begin{table}[!t]
\centering
\caption{An ablation study investigating the independence and contribution of the sub-components within the BIM.}
\label{tab:BIM_dual_path}
\begin{tabular}{lcccc}
\toprule
Configuration & mIoU $\uparrow$ & nIoU $\uparrow$ & Pd $\uparrow$ & Fa $\downarrow$ \\
\midrule
w/o Global Self-Attention & 76.85 & 76.13 & 89.56 & 15.03 \\
w/o Local Self-Attention & 78.41 & 77.92 & 92.47 & 18.72 \\
DENet w/o Gaussian Bias & 82.15 & 78.90 & 97.50 & 13.21 \\
\textbf{DENet} & \textbf{83.96} & \textbf{80.63} & \textbf{98.36} & \textbf{10.93}\\
\bottomrule
\end{tabular}
\end{table}
\noindent\textbf{Influence of BIM.} The Bidirectional  Interaction Module (BIM) is pivotal for DENet's performance, as evidenced by ablation studies. Table \ref{tab:core_components} shows that adding only BIM suppresses background noise ($F_a$: $20.32 \times 10^{-6} \rightarrow 18.82 \times 10^{-6}$) but yields limited edge precision (mIoU: 75.35\%), highlighting its role in semantic alignment rather than detailed boundary recovery. Table \ref{tab:BIM_dual_path} further reveals that the full BIM, which integrates both Local Self-Attentions (preserves edges, Pd: 92.47\%) and Global Self-Attention with Gaussian bias (suppresses clutter, $F_a$: $15.03 \times 10^{-6}$), achieves the optimal balance, boosting mIoU to 83.96\% and reducing $F_a$ to 10.93 $\times$10$^{-6}$. Removing the Gaussian bias causes a significant performance drop (mIoU: -1.81\%, $F_a$: +2.88$\times$10$^{-6}$), confirming its importance in focusing on compact target structures. In summary, BIM enables robust cross-path alignment between semantics and edges, substantially improving shape consistency and false alarm suppression in low-SCR environments.

\noindent\textbf{Role of Multi-ER.} Based on the ablation studies in Table \ref{tab:core_components} and Table \ref{tab2}, the Multi-ER plays a critical role in enhancing edge localization and suppressing false alarms. As shown in Table \ref{tab:core_components}, using Multi-ER alone significantly improves mIoU (77.56\%) and nIoU (76.85\%) compared to the baseline, demonstrating its effectiveness in refining target contours. Furthermore, Table \ref{tab2} reveals that increasing the number of Multi-ER blocks from 0 to 3 leads to progressive gains in all metrics, with the full three-block configuration achieving the best performance (mIoU: 83.96\%, nIoU: 80.63\%, Pd: 98.36\%, Fa: 10.93$\times$10$^{-6}$), confirming that cascaded edge refinement is essential for accurate and robust infrared small target detection under noisy and low-contrast conditions.

\begin{figure*}[t]
	\centering
	\includegraphics[width=\textwidth]{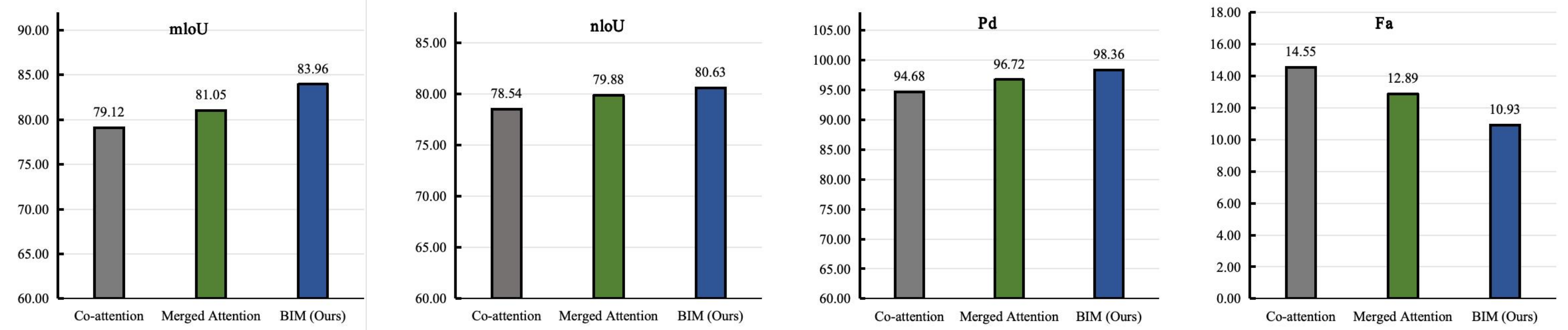}
	\caption{Ablation study results for different fusion strategies within the BIM.
}
    \label{fig_fusion}
\end{figure*}

\begin{table}[t]
  \centering
  \caption{An ablation study on different component combinations within the Multi-ER.}
  \label{tab2}
  \begin{tabular*}{\columnwidth}{@{\extracolsep{\fill}}l cccc}
    \toprule
    Equation Type & mIoU $\uparrow$ & nIoU $\uparrow$ & Pd $\uparrow$ & Fa $\downarrow$ \\
    \midrule
    w/o Gated Conv only   & 76.53 & 76.25 & 95.92 & 14.21\\
    Gated Conv+Bottle Neck& 77.58 & 77.24 & 96.64 & 13.02\\
    Gated Conv+ResBlock   & 78.34 & 78.89 & 97.47 & 12.26\\
    \textbf{DENet}     & \textbf{83.96} & \textbf{80.63} & \textbf{98.36} & \textbf{10.93} \\
    \bottomrule
  \end{tabular*}
\end{table}

\begin{figure*}[t]
	\centering
	\setlength{\tabcolsep}{1pt}
	\scriptsize
	\includegraphics[width=\linewidth]{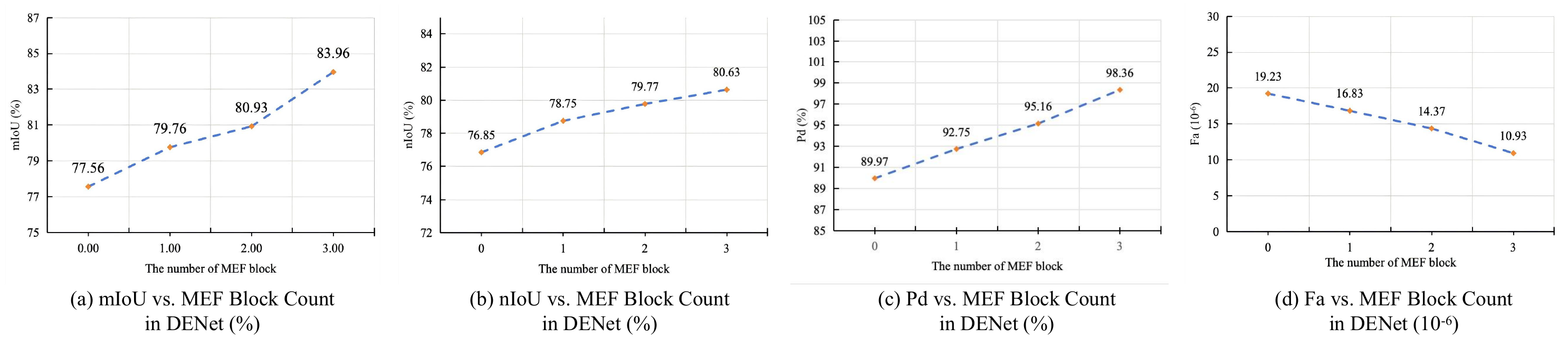}
	\caption{Ablation study on the impact of the number of Multi-Edge Refiner (Multi-ER) blocks on model performance.}
        \label{fig_edge}
\end{figure*}

\subsection{More Results}

\noindent\textbf{Analysis of Fusion Strategies in BIM.} To validate the superiority of our proposed dual-path cross-attention mechanism in BIM, we compared it against two alternative fusion strategies depicted: (a) Co-attention and (b) Merged Attention. As shown in Fig. \ref{fig_fusion}, the Co-attention strategy, which directly concatenates features before self-attention, suffers from feature interference in low-SCR scenarios, yielding a modest mIoU of 79.12\% and a high Fa of $14.55 \times 10^{-6}$. The Merged Attention strategy, which processes features sequentially, leads to the degradation of edge details during deep fusion, achieving an mIoU of 81.05\%. In contrast, our BIM architecture maintains independent feature streams while enabling bidirectional alignment, which preserves fine-grained edge details and leverages global context to suppress clutter. This synergistic design achieves state-of-the-art results with an mIoU of 83.96\% and the lowest Fa of $10.93 \times 10^{-6}$, demonstrating its effectiveness and advanced design.

\begin{figure}[t!]
	\centering
	\includegraphics[width=\linewidth]{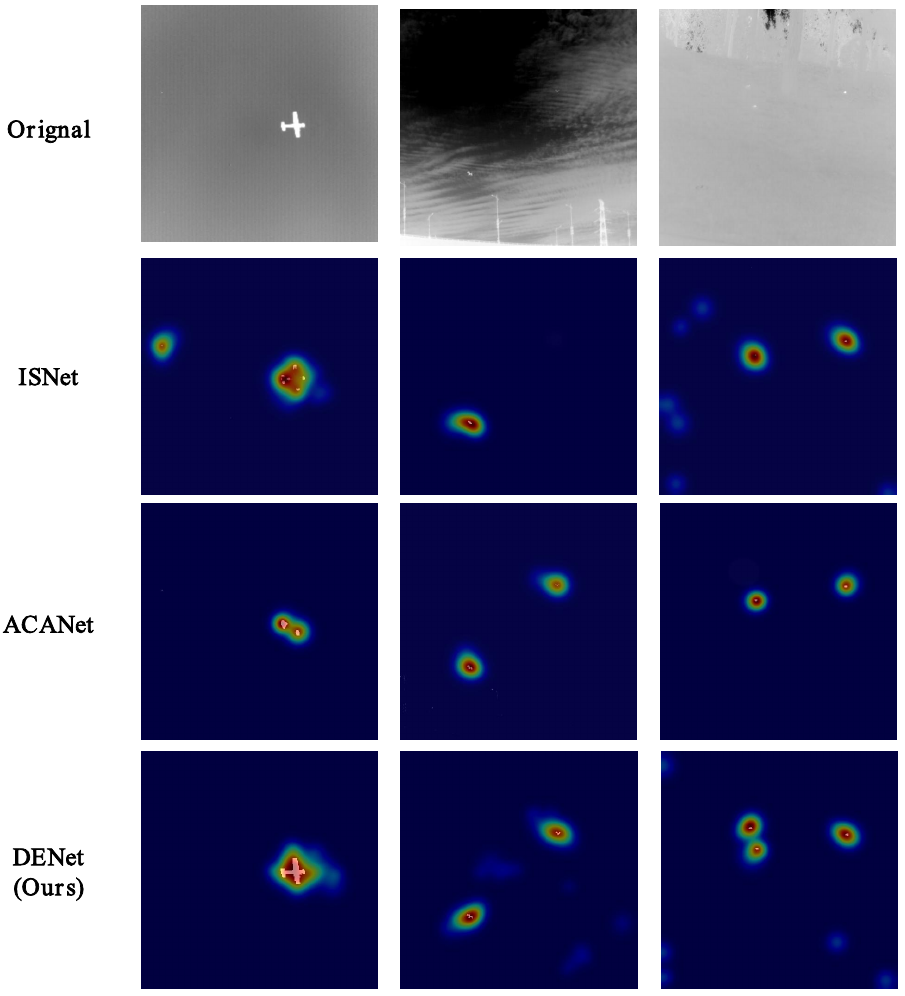}
	\caption{A qualitative visualization comparing the feature maps and detection results.}
    \label{fig_8}
\end{figure}

\begin{table*}[t]
\centering
\caption{Ablation Study on the Universality of the BIM and Multi-ER Modules. This table demonstrates the plug-and-play capability of BIM and Multi-ER by integrating them into different baseline models on the NUDT-SIRST dataset. The consistent performance gains confirm their model-agnostic enhancement capabilities.}
\label{tab:universality}
\begin{tabular*}{\textwidth}{@{\extracolsep{\fill}} l l cccc @{}}
\toprule
Baseline Model & Configuration & mIoU $\uparrow$ & nIoU $\uparrow$ & Pd $\uparrow$ & Fa $\downarrow$ \\
\midrule
\multirow{4}{*}{Unet \cite{b4}} & Original & 74.12 & 72.53 & 91.42 & 18.55 \\
 & + BIM & 77.28 & 76.91 & 93.55 & 15.74 \\
 & + Multi-ER & 78.35 & 77.81 & 94.68 & 14.21 \\
 & \textbf{+ Ours} & \textbf{79.51} & \textbf{78.64} & \textbf{95.12} & \textbf{13.08} \\
\midrule
\multirow{4}{*}{Trans-Unet \cite{b23}} & Original & 80.55 & 82.88 & 97.99 & 11.77 \\
 & + BIM & 81.93 & 83.45 & 98.15 & 11.02 \\
 & + Multi-ER & 82.14 & 84.05 & 98.21 & 10.53 \\
 & \textbf{+ Ours} & \textbf{82.76} & \textbf{84.31} & \textbf{98.29} & \textbf{10.17} \\
\midrule
\multirow{2}{*}{Swin-Unet \cite{b5}} & Original & 75.35 & 75.21 & 88.14 & 18.82 \\
 &+ BIM         & 75.35 & 75.21 & 88.14 & 18.82\\
 &+ Multi-ER    & 77.56 & 76.85 & 89.97 & 19.23\\
 & \textbf{+ Ours} & \textbf{83.96} & \textbf{80.63} & \textbf{98.36} & \textbf{10.93} \\
\bottomrule
\end{tabular*}
\end{table*}
\noindent\textbf{Analysis of Injection Location in Multi-ER.} Based on Fig. \ref{fig_edge} (Ablation study on the different number of Multi-ER blocks), the analysis of the injection location within the Multi-Edge Refiner (Multi-ER) reveals that increasing the number of Multi-ER blocks significantly enhances model performance. As the number of blocks increases from 0 to 3, all key performance metrics show marked improvement: the mIoU increases from 75.35\% to 83.96\%, the nIoU from 75.21\% to 80.63\%, and the Probability of Detection (Pd) surges from 88.14\% to 98.36\%.  Concurrently, the False Alarm Rate (Fa) is substantially reduced from $18.82 \times 10^{-6}$ to $10.93 \times 10^{-6}$. This indicates that cascaded Multi-ER blocks are crucial for progressively refining edge features, thereby improving the accuracy and reliability of target detection.

\noindent\textbf{Visualization of the Feature Map.} As illustrated in Fig. \ref{fig_8}, we provide a qualitative comparison of detection results to visualize the model's effectiveness. The figure displays the original infrared image with a low-contrast target, the baseline model's detection result, and the result from our DENet model. The baseline model, while able to roughly locate the target, produces a diffuse and imprecise heatmap with significant background noise, indicating poor feature representation and an inability to clearly distinguish the target from its surroundings. In stark contrast, the feature map generated by DENet shows a sharply focused, high-intensity activation precisely on the target area, with surrounding background clutter almost entirely suppressed. This demonstrates DENet's superior feature learning capability, where the Multi-ER and BIM modules work synergistically to enhance target features while filtering out interference, resulting in a clean and accurate feature representation that leads to precise localization and high confidence in detection.

\noindent\textbf{Universality.} The universality of our proposed modules is further validated in Table~\ref{tab:universality}, where both the Bidirectional Interaction Module (BIM) and the Multi-Edge Refiner (Multi-ER) exhibit strong plug-and-play capabilities across diverse backbone architectures. When integrated into models such as Swin-Unet \cite{b5} and MSHNet \cite{b42}, both modules consistently enhance performance metrics including mIoU, nIoU, Pd, and Fa, underscoring their architectural independence and broad applicability. This model-agnostic improvement stems from the complementary strengths of BIM in semantic-edge alignment and Multi-ER in fine-grained edge recovery, which address fundamental challenges in infrared small target detection irrespective of the base network. Their synergistic combination not only boosts detection accuracy and shape consistency but also significantly reduces false alarms, confirming their versatility and potential as standard components for advancing IRSTD systems.

\section{CONCLUSION}
The DENet framework has revolutionized infrared dim-small target detection by pioneering the synergistic evolution of differentiable edge modeling and semantic awareness, which breaks the century-long limitation of traditional fixed gradient operators through an adaptive second-order differential operator. Its innovative dual-path architecture establishes dynamic equilibrium between noise suppression and edge fidelity. The Multi-ER reconstructs edge response mechanisms via parameterized differential equations to significantly enhance robustness in low-contrast scenarios, while the feature alignment module constructs a cross-modal attentional dialogue mechanism enabling spatial alignment between local texture details and global semantic context. Under complex background interference, this framework achieves a breakthrough Boundary IoU accuracy of 68.33\% while reducing the false alarm rate to 13.61$\times$10$^{-6}$, establishing a new technical paradigm for high-reliability remote sensing monitoring.

\begin{IEEEbiography}[{\includegraphics[width=1in,height=1.25in,clip,keepaspectratio]{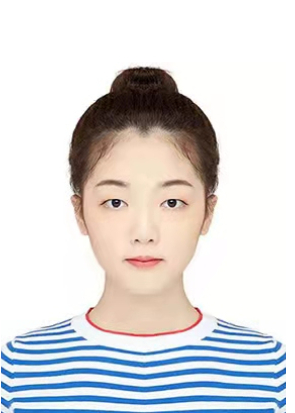}}]{Jiayi Zuo}
received the B.Eng. degree in E-Commerce and Law from Beijing University of Posts and Telecommunications (BUPT), Beijing, China, in 2024. She is currently pursuing the M.S. degree in Information and Communication Engineering at BUPT, focusing on remote sensing, sensor fusion, and object detection.
\end{IEEEbiography}

\begin{IEEEbiography}[{\includegraphics[width=1in,height=1.25in,clip,keepaspectratio]{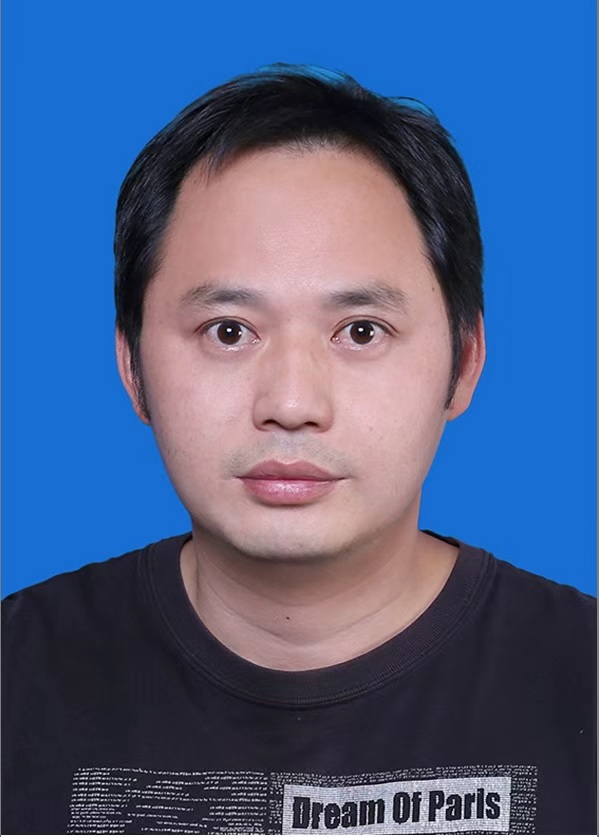}}]{Songwei Pei} received his Ph.D. degree in computer science from the Institute of Computing Technology (ICT), Chinese Academy of Sciences (CAS), Beijing, China, in 2011. He is currently an Associate Professor with the School of Computer Science (National Pilot Software Engineering School), Beijing University of Posts and Telecommunications, Beijing, China. He has published several dozen technical papers in renowned international journals and conferences, including the IEEE/ACM Transactions series, Elsevier and Springer journals, and others. He serves as a reviewer for several renowned journals and as a program committee member for significant conferences. Additionally, he is a member of several academic and specialized committees. His current research interests include computer vision, deep learning, computer architecture, and VLSI design.\end{IEEEbiography}

\begin{IEEEbiography}
[{\includegraphics[width=1in,height=1.25in,clip,keepaspectratio]{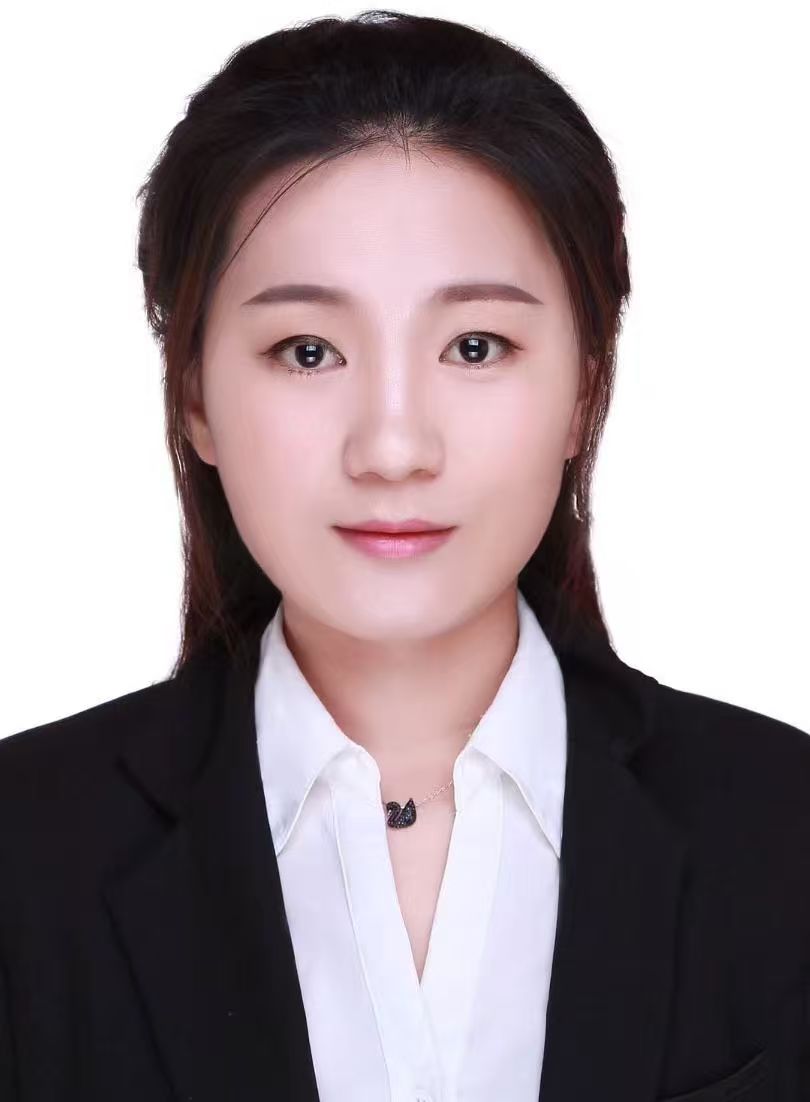}}]{Qian Li} is currently an Assistant Professor at Beijing University of Posts and Telecommunications (BUPT). Her research interests focus on natural language processing, multimodal learning, and knowledge graphs. She has authored over 30 papers published in top-tier conferences and journals such as NeurIPS, WWW, ACM MM, and AAAI, amassing over 2,000 citations on Google Scholar. She received the Best Paper Nomination at CIKM 2022, two ESI Highly Cited Papers, IEEE TCCLD 2024 Technical Innovation Award, the CodaLab championship, and the First Prize for Scientific and Technological Progress from the State Grid Big Data Center. She has also served as an Area Chair for ACL and NeurIPS, a Senior Program Committee member for IJCAI, and a Program Committee member for TKDE, ICML, NeurIPS, and ICLR.
\end{IEEEbiography}

\begin{IEEEbiography}[{\includegraphics[width=1in,height=1.25in,clip,keepaspectratio]{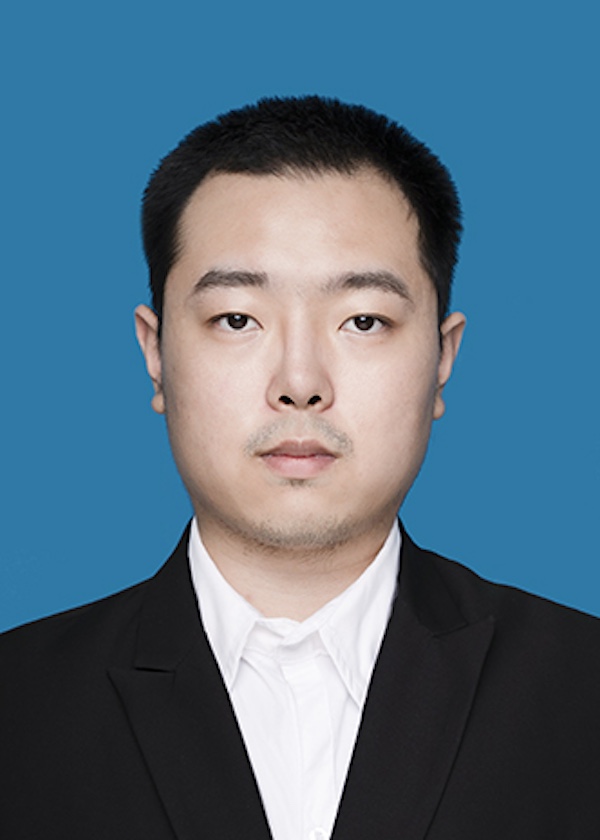}}]{Yuanzhou Huang} received the B.Eng. degree in software engineering from the School of Information Science and Engineering, Yanshan University, Hebei, China, in 2021.   He is currently working toward the Ph.D. degree in computer science with the School of Computer Science (National Pilot Software Engineering School), Beijing University of Posts and Telecommunications, Beijing, China. His current research interests include object detection and multi-object tracking.\end{IEEEbiography}

\begin{IEEEbiography}
[{\includegraphics[width=1in,height=1.25in,clip,keepaspectratio]{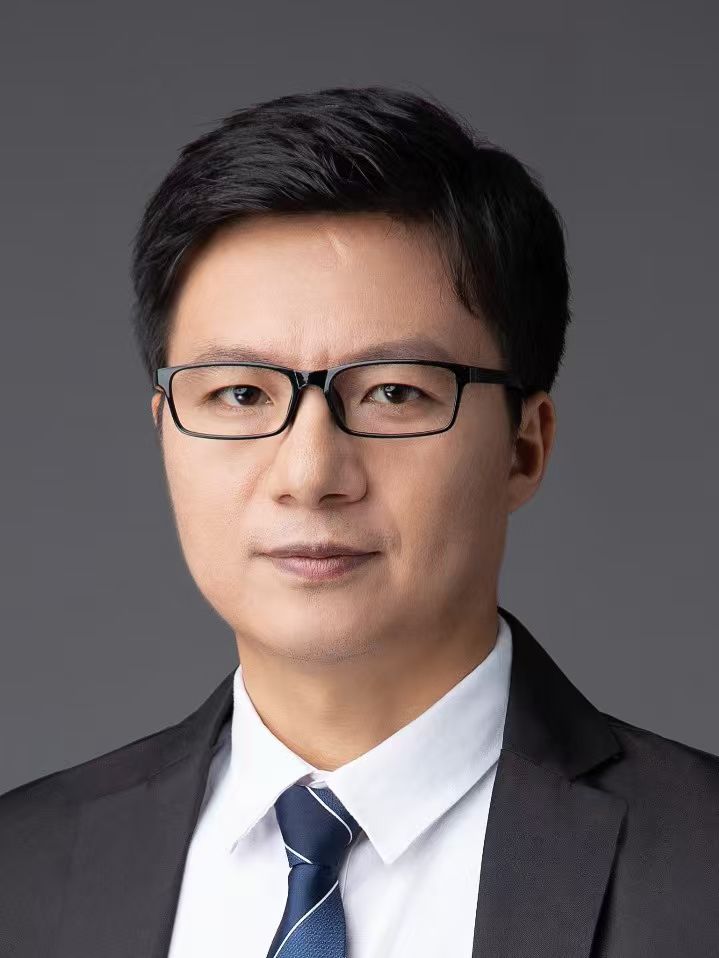}}]{Shangguang Wang}(Senior Member, IEEE) received the PhD degree from the Beijing University of Posts and Telecommunications, Beijing, China, in 2011. He is a professor with  the School of Computer Science (National Pilot School of Software Engineering), Beijing University of Posts and Telecommunications, China. He has published more than 150 papers. His research interests include service computing, mobile edge computing, and satellite computing. He is currently serving as chair of the IEEE Technical Committee on Services Computing, and vice-chair of the IEEE Technical Committee on Cloud Computing. He also served as general chairs or program chairs of more than ten IEEE conferences, and associate editors of several journals, such as IEEE Transactions on Services Computing, Journal of Software: Practice and Experience, and so on. He is a fellow of the IET.
\end{IEEEbiography}
\bibliographystyle{IEEEtran}
\bibliography{ref.bib}

\vspace{11pt}

\vfill

\end{document}